\documentclass[12pt,twoside,a4paper]{irs}
\usepackage{epsf,exscale,times}
\usepackage[english]{babel}
\usepackage{graphicx}
\usepackage{amsmath}
\usepackage{fancyhdr}

\newcommand {\vsp}   {\vspace*}

\def\title#1{\vsp{-16mm}\begin{center}\Large\bf{#1}\end{center}\vsp{0mm}}
\def\author#1{{\begin{center}\textbf{#1}\end{center}\vspace{-1mm}}}
\def\address#1{\vsp{-3mm}\begin{center}\baselineskip12pt\normalsize{#1}\end{center}\vsp{-1mm}}

\def\abstract#1{{\vspace{-5mm}
    \begin{center}
      \begin{minipage}{0.85\textwidth}
        \noindent\bf \textit{Abstract:}
        \small\rm\emph{#1}
				\vsp{-0.5em}
      \end{minipage}
    \end{center}
}}

\def\authorsheadline=#1{\global\def\@authorsheadline{#1}}
\global\def\@authorsheadline{}

\def\TeX{T\kern-.1667em\lower.5ex\hbox{E}\kern-.125emX}

\def\LaTeXG{{\rm L\kern-.36em\raise.3ex\hbox{\sc a}\kern-.15emT\kern-.1667em\lower.7ex\hbox{E}\kern-.125emX}}
\def\LaTeXK{{\it L\kern-.24em\raise.4ex\hbox{\scriptsize \it A}\kern-.20emT\kern-.1667em\lower.5ex\hbox{E}\kern-.125emX}}

\begin{document}

\fancypagestyle{firststyle}
{
   \fancyhf{}
   \lfoot{ \footnotesize{The International Radar Symposium IRS 2025, May 21-23, 2025, Berlin\break     ISBN 978-3-00-079305-9 \copyright  2024 DGON} }
   \rfoot{ \footnotesize {\thepage} }
}

\thispagestyle{firststyle}
\fancyhf{}
\renewcommand{\headrulewidth}{0pt}
\renewcommand{\footrulewidth}{1pt}
\renewcommand{\footskip}{50pt}

\pagestyle{fancy}
\fancyfoot[RO,LE]{ \footnotesize {\thepage} }

\title{Enhancing Fourier-based Doppler Resolution \\ with Diffusion Models}


\author{
Denisa Qosja, Kilian Barth, Simon Wagner}
\address{
    Cognitive Radar Methods, Fraunhofer FHR\\
    Wachtberg, Germany\\
	 \{denisa.qosja, kilian.barth, simon.wagner\}@fhr.fraunhofer.de
		}

\abstract{
In radar systems, high resolution in the Doppler dimension is important for detecting slow-moving targets as it allows for more distinct separation between these targets and clutter, or stationary objects. However, achieving sufficient resolution is constrained by hardware capabilities and physical factors, leading to 
the development of processing techniques to enhance the resolution after acquisition. In this work, we leverage artificial intelligence to increase the Doppler resolution in range-Doppler maps. Based on a zero-padded FFT, a refinement via the generative neural networks of diffusion models is achieved. We demonstrate that our method overcomes the limitations of traditional FFT, generating data where closely spaced targets are effectively separated. 
}



\section{Introduction}

The resolution of a radar system defines its ability to distinguish between  target returns in presence of clutter~\cite{skolnik_book, radar_resolution}. Slow-moving targets pose recognition challenges, given that they typically produce near-zero Doppler shifts and thus might blend in with present clutter. In scenarios where only moving targets are of interest, returns that produce near zero Doppler might be filtered out. The subtle Doppler shifts of these targets require high resolution, which allows the separation of targets from background interference, and enhances the signal-to-clutter ratio.

Achieving higher resolution in Doppler is cumbersome. Integration time, a key limiting factors, increases the processing complexity and restricts real-time applicability. Over the years, efforts have been directed towards developing algorithms to actually enhance the resolution of data. A common and traditional approach is Fast Fourier Transform (FFT) combined with zero-padding. Zero-padding increases the resolution of the FFT by artificially extending the length of the signal, effectively increasing the number of frequency bins. However, this method is still constrained by the length of the observation period and may not fully resolve small shifts. Time frequency analysis methods, such as STFT and spectral estimation techniques were subsequently developed to overcome the limitations of FFT. The latter focus on detecting the frequency components of the signal with higher resolution, even when the frequency shifts are very small and close together. These techniques usually analyze the signal’s covariance matrix and decompose it into components, effectively enhancing the resolution beyond what is achievable through basic zero-padding. MUltiple SIgnal Classification (MUSIC)~\cite{MUSIC_orig, radar_MUSIC} and Estimation of Signal Parameters by Rotational Invariance Techniques (ESPRIT)~\cite{ESPRIT_orig} are common spectral estimation algorithms used in radar~\cite{degraaf_sar_spectral_estimation}. They yield adequate performance in comparison to other methods, however they are limited by their inability to capture complex patterns in the data, particularly in challenging environments where conditions may be dynamic or cluttered.

 
Learning-based algorithms, specially deep neural networks (NN) have been explored to better capture the intricate and non-linear relationships in the data.
Convolutional Neural Networks (CNN) are among the first learning-based methods that tackled the resolution problem in 2D data. They have shown to be effective yet limited in their ability to produce high-quality and realistic outputs, especially when dealing with complex structures or noisy environments. Generative models such as Generative Adversarial Networks (GANs)~\cite{gan_goodfellow} or Diffusion Models (DMs)~\cite{ddpm} are convolutional-based networks that learn the underlying distribution of the data, capturing the convoluted relationships structures, and variations in the data. They can better handle the ambiguity and uncertainty inherent in SR tasks, especially when large amounts of information are missing. 
Diffusion Models are the state-of-the-art generative models, achieving high quality and fidelity in data. In this work, we exploit a DM-based network, namely Super-Resolution via Repeated Refinement SR3~\cite{SR3} for image super-resolution in range-Doppler (RD) maps.

\section{Super-Resolution with Diffusion Models}

Super Resolution (SR) is an inverse problem, that involves the reconstruction of high-resolution (HR) data from low-resolution (LR) observations which do not include or might have lost the information on high-frequency details. Mathematically, this problem can be modeled by a degradation function:
\begin{equation}
	\mathbf{x}_{LR} = D(\mathbf{x}_{HR}, \mathbf{\Theta}) = (\mathbf{x}_{HR} \otimes \mathbf{k})\downarrow_s + \mathbf{n}, 
\end{equation}

where $D$ is the degradation function and $\mathbf{\Theta}$ consists of degradation parameters, such as blur $\mathbf{k}$, noise $\mathbf{n}$ and downsampling factor $s$. Since the degradation is typically unknown, the main challenge is determining the inverse mapping $D^{-1}$, with parameters $\mathbf{\theta}$ that are usually embodied as a SR model. The inverse recovery of $\mathbf{x}_{HR}$ from $\mathbf{x}_{LR}$, $\mathbf{x}_{SR} = D^{-1}(\mathbf{x}_{LR})$, is underdetermined as the degradation process removes information, leaving multiple plausible HR data that could correspond to the same LR observation. Consequently, SR is a challenging problem that requires robust methods to restore lost details and remove noise. 
\begin{figure}[hbt!]
	\centering
	\includegraphics[width=1\textwidth]{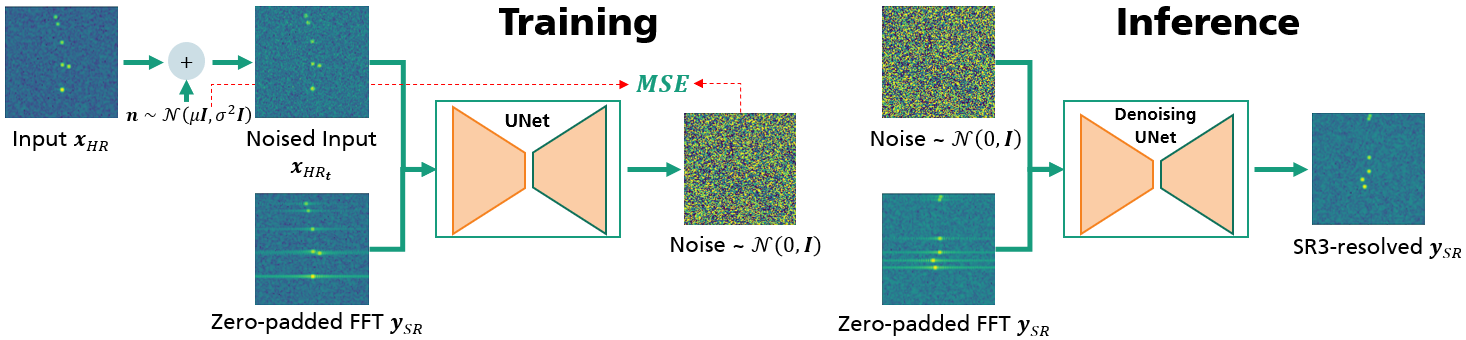}
	\
	\caption{Training and inference for SR3. During training, RD maps are infused with noise, concatenated with the corresponding zero-padded FFT maps and fed to the UNet, which learns to predict the noise added to the input samples. During inference, noise sampled from a normal distribution is concatenated with a chosen RD map and fed to UNet that iteratively removes the artifacts in order to obtain the clean RD map.}
	\label{fig:teaser_fig}
\end{figure} 

\subsection{SR via Repeated Refinement (SR3)}

The inverse function in SR3 is designed as a Denoising Diffusion Probabilistic Model (DDPM)~\cite{ddpm}, where data is initially resolved via traditional techniques, then refined and translated to high-frequency details by the network. By using the zero-padded FFT as input, the SR3 is already prepared to have the same spatial dimensionality as the target data. This reduces the need to employ specific layers in the network to address the upsampling of the data. Originally, this network was designed for RGB data, which are preprocessed with bicubic interpolation to increase the resolution and align the number of samples.  
To adapt the network more closely to the physics domain, our work replaces the interpolation technique by FFT with zero-padding. 

Given a dataset of data pairs between SR maps $\mathbf y_{SR}$, obtained from the zero-padded FFT, and HR maps $\mathbf x_{HR}$, the samples are assumed to be drawn by an unknown conditional distribution $p(\mathbf{y}_{SR}, \mathbf{x}_{HR})$. Designed as a conditional image generation model, SR3 learns a parametric approximation to $p(\mathbf{x}_{HR} |\mathbf{y}_{SR})$ through a stochastic iterative refinement process that maps source data $\mathbf{y}_{SR}$ to the target $\mathbf{x}_{HR}$.
The training phase of the network is characterized by a forward diffusion process, where Gaussian noise is fused with the HR maps for $T$ diffusion steps. The diffusion follows a Markov chain with transition probabilities $q(\mathbf x_{{HR}_t} | \mathbf x_{{HR}_{t-1}}, \mathbf y_{SR})$, generating a latent variable at the latest diffusion step. Due to the specific choice of the parameters for SR3, the latent variable is obtained as pure noise $ \sim \mathcal{N}(\mathbf 0, \mathbf I)$. During the inference, noise sampled from standard normal distribution is concatenated with $\mathbf{y}_{SR}$ and given as input to the model. The model iteratively refines the data through successive iterations according to learned conditional transition distributions  $p_{\theta}(\mathbf x_{{SR}_{t-1}} | \mathbf x_{{SR}_t}, \mathbf y_{SR})$ to eventually generate, at diffusion time $t=0$, the super-resolved map $\mathbf x_{SR}$, as it was sampled from $p(\mathbf{x}_{HR} |\mathbf{y}_{SR})$.  Similar to DDPM, SR3 is constrained to minimize the difference between distribution $p$ and $q$ via Kullback-Leibler divergence. The training, as well as the inference process of SR3 are shown in Fig.~\ref{fig:teaser_fig}.

\vspace{-0.2cm}
\subsection{Model Architecture}

The original SR3 network replaces the original DDPM residual blocks with residual blocks from BigGAN~\cite{biggan}, and adds an additional scaling to the skip connections by $\frac{1}{\sqrt{2}}$. Moreover, the number of residual blocks is increased and the channel multipliers at different resolutions are set to $[1, 2, 4, 8, 8]$. However, we notice that for our dataset this architecture generates data abundantly prone to color shifting. This issue is attributed to measures taken to reduce the computational costs, such as lowering the batch size or decreasing the learning period. To avoid this problem, the network we use to train with RD maps is completely based on the architecture of DDPM~\cite{qosja_ddpm}, given that it is less computationally expensive. The initial inner channel is set to $64$, and increased with multiplying factors $[1, 2, 3, 4]$.  
Regarding the design choice, the diffusion steps are set to $T = 2000$, while the range of the variance $\beta$ is set differently to the DDPM model with $\beta_1= 10^{-6}$ and $\beta_T=0.02$. 

\begin{figure}[t!]
	\centering
	\includegraphics[width=1\textwidth]{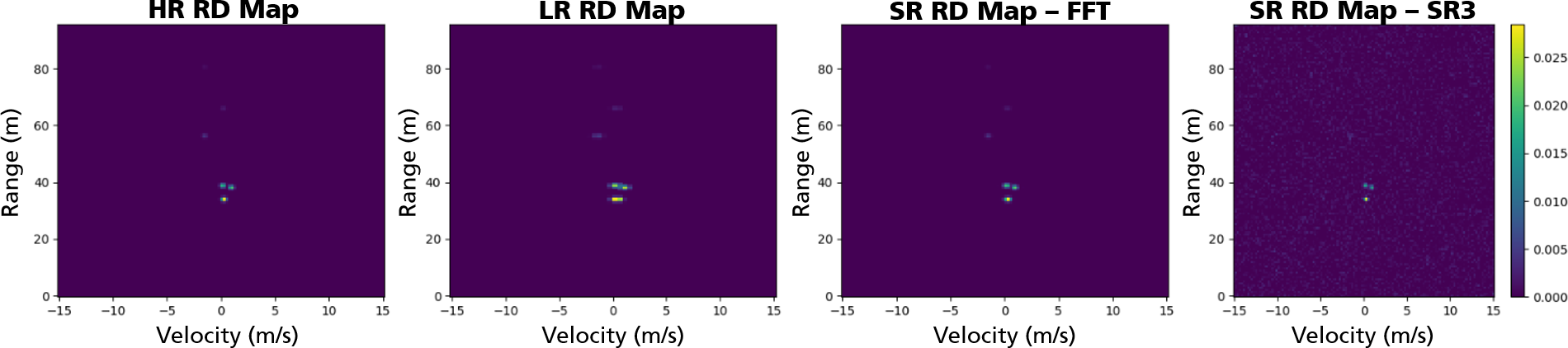}
	\caption{HR map, LR map, SR map via zero-padded FFT and SR map via SR3 in linear scale given from left to right. RD maps are resolved from a LR version of downsampling factor of $2$.}
	\label{fig:exp_1_sr64}
\end{figure} 

\vspace{-0.2cm}
\section{Dataset}

We exploit simulated RD maps for super-resolved Doppler. In this work, data with three and ten point targets with certain velocity are generated through a simulation pipeline, captured by two ULA elements separated in distance by $0.12$m. Additionally, clutter points with near-zero velocity are introduced. The number of range and Doppler bins for the HR maps are set to $128$x$128$, while the LR version for each map is obtained by reducing the integration time by factors of $2$, $4$, and $8$. A maximum unambiguous range of $96$m is set, with range resolution of $0.75$m and velocity resolution $0.23$m/s. A conventional super-resolution (SR) version of the RD maps has been obtained via FFT with zero-padding. 
Examples of RD maps are demonstrated in Fig.~\ref{fig:exp_1_sr64}, where the three leftmost maps correspond, from left to right, to the HR map, followed by the LR map, and then the SR map. The maps contain only three targets and thee clutter points and they are shown here in linear scale.

Before being fed to the network, the maps are processed. Firstly, they are filtered with a tapering Blackman window, and then normalized to range $[-1, 1]$, as required by the diffusion network. 

\vspace{-0.2cm}
\section{Experiments}

\begin{figure}[t!]
	\centering
	\includegraphics[width=1\textwidth]{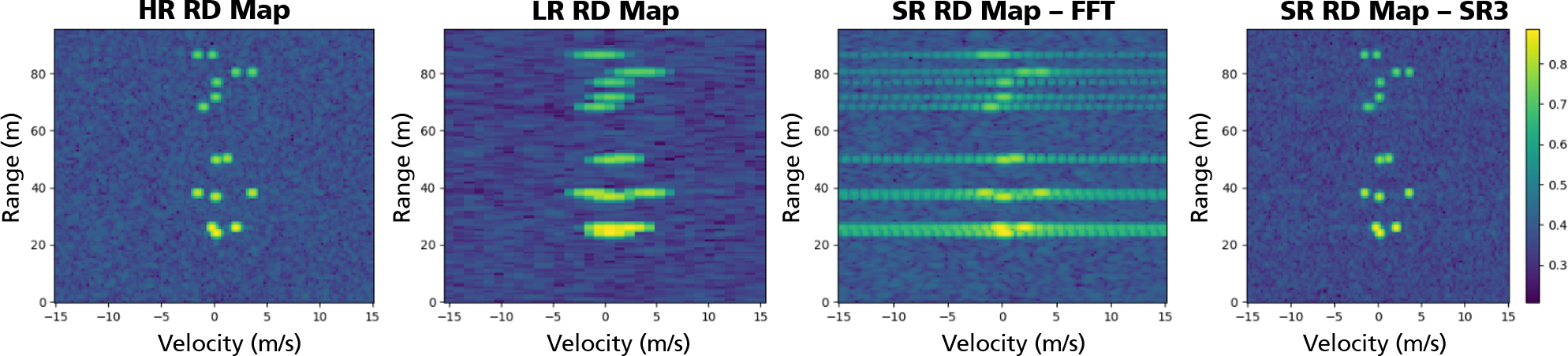}
	\caption{HR map, LR map, SR map via zero-padded FFT and SR map via SR3 in logarithmic scale given from left to right. RD maps are resolved from a LR version of downsampling factor of $4$.}
	\label{fig:exp_2_sr32_1}
\end{figure}

In this section, the effectiveness of the super-resolution network is assesed on the introduced dataset. 
We use GPU cards A$100$ with $80$GB, which help to accommodate a batch size of $128$. 
Firstly, experiments with downsampling factor of $2$ are conducted. The RD maps fed to the network are set simply in linear scale. For this experiment, the number of epochs is increased to $300$, due to the continous downward behaviour of the training loss curve. Fig.~\ref{fig:exp_1_sr64} shows the HR RD map, the LR version and the SR output obtained from the trained network. As expected, in the LR map, the second and the third closest targets are no longer separable. In the SR map obtained from SR3, these targets can be distinguished, however it is also noticable that the noise level is much higher than the original map. As a consequence, the furthest targets are immersed in noise and will not be detected by CFAR. Since the number of epochs is considerably high, increasing it further would not be a meaningful solution.  
Therefore, we input the data into the network directly in logarithmic scale. The number of epochs is set only to $100$, as we notice that the network is able to denoise and reconstruct the maps in less training time. In this experiment, the output from SR3 is less noisy and the furtherst targets are easily detectable. In Fig.~\ref{fig:exp_2_sr32_1}
, results for downsampling factor of $4$ using RD maps with ten targets and five clutter points are shown.

\begin{figure}[t!]
	\centering
	\includegraphics[width=1\textwidth]{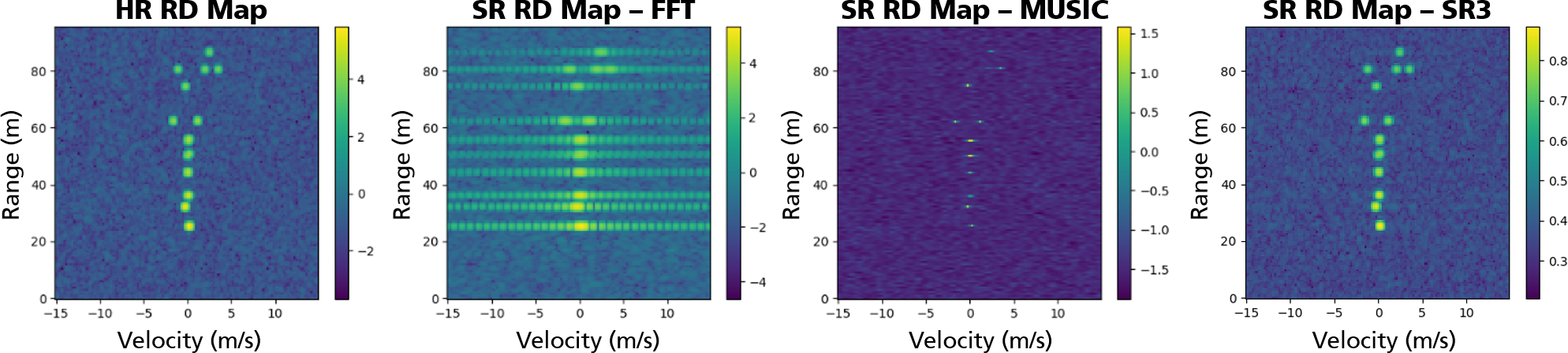}
	\caption{HR (left), SR FFT (middle), and SR3 (right) range-Doppler maps with downsampling factor of $8$ in logarithmic scale.}
	\label{fig:exp_3_sr4_methods}
\end{figure}

\begin{figure}[t!]
	\centering
	\includegraphics[width=1\textwidth]{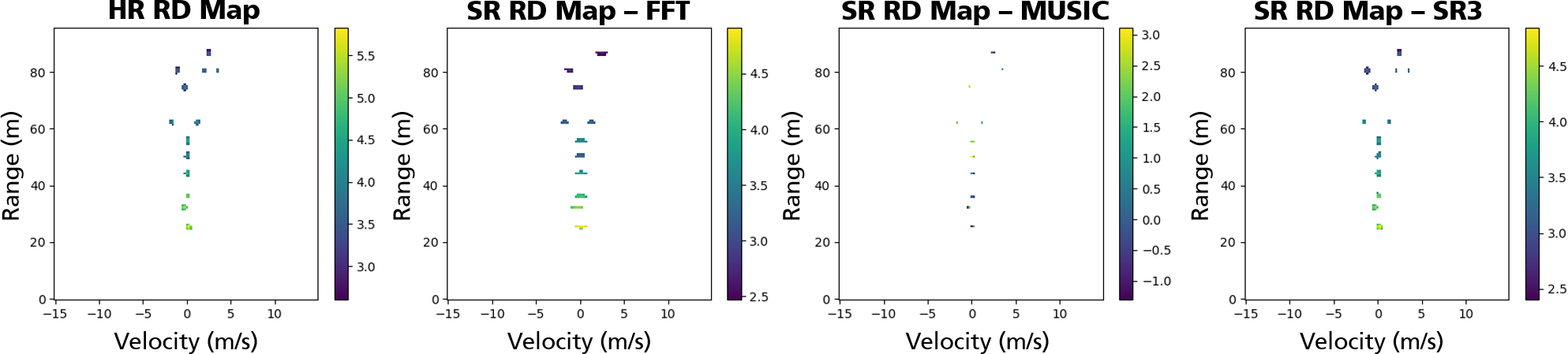}
	\caption{HR map, and CFAR detections from the SR map via zero-padded FFT, the SR map via MUSIC and the SR map via SR3 given from left to right. RD maps are resolved from a LR version of downsampling factor of $4$.}
	\label{fig:exp_4_cfar_detections}
\end{figure}

Furthermore, we compare the outcome of the network with the result of MUSIC algorithm, and use CFAR for detecting the targets. 
In Fig.~\ref{fig:exp_3_sr4_methods}, the super-resolved RD Maps for another scenario are compared, including the results from MUSIC. At the range close to 80m, the two left most targets are dimly reconstructed with MUSIC, while with FFT the two rightmost targets are not distinguishable due to the ripple artifacts. We feed all maps to the CFAR detector and show the outcomes in Fig.~\ref{fig:exp_4_cfar_detections}. As expected, the detections of the RD map resolved from zero-padded FFT are not completely accurate, with two targets being undetected. Similarly, the observed output of CFAR on the RD map resolved with MUSIC, indicates that two targets are not distinguished. This means that MUSIC lacks to accurately resolve the LR map with a downsampling factor of $4$. Meanwhile, CFAR applied on the SR3 map is able to correctly detect all the targets that are simulated.

\begin{figure}[t!]
	\centering
	\includegraphics[width=\textwidth]{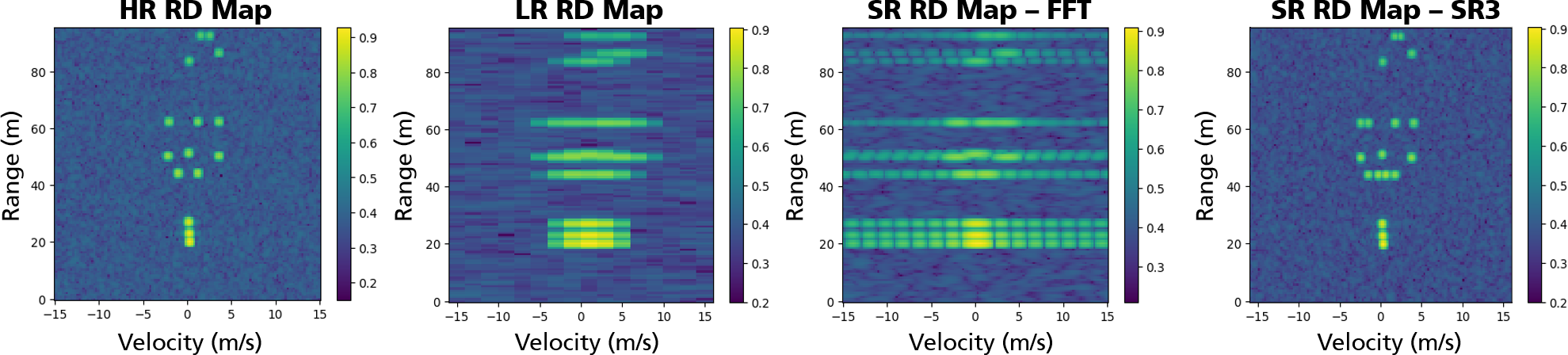}
	\caption{HR (left), SR FFT (middle), and SR3 (right) range-Doppler maps with downsampling factor of $8$ in logarithmic scale.}
	\label{fig:exp_5_sr8}
\end{figure}

To asses the limitations of SR3 for this dataset, experiments with a downsampling factor of $8$ are conducted. Accurate super-resolving is noticed in various RD maps, however, in Fig.~\ref{fig:exp_5_sr8} a failure case of SR3 is visualized. Extra targets have been artificially created at range near $40m$, which shows the presence of hallucinating effects that generative networks experience.

\vspace{-0.2cm}
\section{Conclusion}

We introduce a learning-based method to serve as a filter for increasing the Doppler resolution before performing the zero-Doppler filter for the separation of moving targets from clutter. The zero-padded FFT is improved by a generative network, namely SR3, which is able to effectively differentiate closely spaced targets. For future work, we aim to further develop this method with measured data, containing more challenging issues and thus introducing a more complex scenario.

\vspace{-0.3cm}

\def\bibname{LIT}

\end{document}